# Optimizing and Fine-tuning Large Language Model for Urban Renewal

Xi Wang[*,1,2]    Xianyao Ling[2]    Tom Zhang[3]    Xuecao Li[4]    Shaolan Wang[1]
Zhixing Li[5]    Liang Zhang[6]    Peng Gong[7]

[1] Tsinghua University    [2] Cross-strait Tsinghua Research Institute
[3] Biocloo Inc    [4] China Agricultural University    [5] Zhipu AI
[6] Beijing Rongchuang Geotechnical Engineering Co., Ltd.    [7] The University of Hong Kong

xi-wang19@mails.tsinghua.edu.cn    xianyao.ling@ctri.org.cn    tzhang@biocloo.com.cn
xuecaoli@cau.edu.cn    sl.wang21@mail.tsinghua.edu.cn    zhixing.li@zhipuai.cn
liangzh@rcgeotec.com    penggong@hku.hk

**Abstract:** This study aims to innovatively explore adaptive applications of large language models (LLM) in urban renewal. It also aims to improve its performance and text generation quality for knowledge question-answering (QA) tasks. Based on the ChatGLM, we automatically generate QA datasets using urban renewal scientific literature corpora in a self-instruct manner and then conduct joint fine-tuning training on the model using the Prefix and LoRA fine-tuning methods to create an LLM for urban renewal. By guiding the LLM to automatically generate QA data based on prompt words and given text, it is possible to quickly obtain datasets in the urban renewal field and provide data support for the fine-tuning training of LLMs. The experimental results show that the joint fine-tuning training method proposed in this study can significantly improve the performance of LLM on the QA tasks. Compared with LoRA fine-tuning, the method improves the Bleu and Rouge metrics on the test by about 5%; compared with the model before fine-tuning, the method improves the Bleu and Rouge metrics by about 15%-20%. This study demonstrates the effectiveness and superiority of the joint fine-tuning method using Prefix and LoRA for ChatGLM in the urban renewal knowledge QA tasks. It provides a new approach for fine-tuning LLMs on urban renewal-related tasks.

1. Introduction

  With the acceleration of global urbanization, urban renewal has become an important research area in urban planning and development. Urban renewal involves optimizing and reshaping urban space, including social, economic, cultural, and other aspects, making it a complex system engineering task. In the process, applying big data and artificial intelligence (AI) technologies becomes particularly important. Among them, large language models, as an essential branch of AI, have broad application prospects in urban renewal.

  A large language model (LLM) is based on deep learning and neural networks for natural language processing (NLP), which provides powerful text generation and comprehension capabilities. In this paper, we aim to explore the application of LLM in urban renewal and improve performance and text generation quality in knowledge question-answering (QA) tasks. The study is based on the ChatGLM [18], and up-to-date urban renewal literature to construct the urban renewal LLM. The main contributions of our paper are as follows:

- Based on urban renewal scientific literature corpora, we write appropriate prompts and guide the ChatGLM to automatically generate QA data in a self-instruct way [41] according to the prompt and given text corpus.
- With the QA instruction dataset, we propose a joint fine-tuning method for ChatGLM, which combines Prefix [34] and LoRA [36] methods to construct the urban renewal LLM, and provides a new approach for fine-tuning LLMs.
- We use Bleu and Rouge metrics to evaluate the model performance of different fine-tuning methods and demonstrate that the proposed joint fine-tuning method can effectively improve the model's performance and text generation quality in urban renewal knowledge QA tasks.

This study is drawn upon various research works to conduct in-depth discussions on applying LLM in urban renewal. By fine-tuning the ChatGLM, we build an urban renewal LLM that provides systematic knowledge on urban renewal tasks, which are essential for urbanization development and digital city applications.

---

[*] Corresponding author.

## 2. Related Work

There have been many related research works in urban renewal and urbanization redevelopment. Zhou et al. propose a new approach that utilizes nighttime light (NTL) observations to develop consistent global urban maps from 1992 to 2013. Their study enables the analysis of urban dynamics and provides essential information for sustainable urban development [1]. Gong reviews the use of remote sensing in monitoring environmental change in China and emphasizes the need for methodological advancements and expanded application fields [2]. Angel et al. examine global urban expansion and project significant increases in urban land cover in developing countries, highlighting the importance of adequate preparations for accommodating this expansion [3]. Wang et al. utilize satellite remote sensing data to analyze the exponential urban expansion in China from 1990 to 2010, focusing on the conversion of croplands into urban areas and its implications for urban planning and management [4]. Gong et al. present the first global land cover mapping results at finer resolutions using Landsat TM and ETM+ data, achieved through the classification of over 8,900 scenes employing various classifiers and a unique land-cover classification system [5]. Chen et al. propose a pixel-object-knowledge-based operational approach for global land cover mapping at 30 m resolution, achieving high overall classification accuracy [6]. Besides, Angel et al. [7], Bao et al. [8], Du et al. [9], Chen et al. [10], and Wang et al. [11] have also contributed to the understanding of global urban dynamics, the utilization of remote sensing data, and the mapping of land cover at various resolutions.

Modeling techniques and the relationships between urban renewal and development have also been explored. Li et al. propose a model that integrates constrained cellular automata and GIS to simulate sustainable urban development, aiming to find better urban forms for sustainable planning [12]. Li et al. propose an ensemble-urban cellular automata model integrated with an uncertainty map, CART, and ANN to improve the performance of a single CA model for urban growth simulation [13]. Li et al. investigate the changing dynamics of equalizing public services in China and its correlation with regional economic disparities, emphasizing the importance of addressing social equalization for long-term regional development [14]. Pendlebury et al. explore the relationship between heritage, urban regeneration, and place-making, highlighting the motivations and dynamics of using heritage in urban development processes worldwide [15]. The actual changes in the urban planning system in cities like Prague, have also been studied to understand the transition to new institutional methodologies [16]. The diverse rationalities in urban design and their implications for sustainable urban development have been analyzed using Cultural Theory [17]. This research provides valuable information for urban renewal LLM creations and improvements.

The success of LLMs in NLP tasks has motivated researchers to explore their potential in urban renewal and development. The General Language Model (GLM) incorporates autoregressive blank infilling and 2D positional encodings [18]. GLM performs well compared to models like BERT [19], T5 [20], and GPT [21] on a wide range of language tasks. The Transformer architecture introduced by Vaswani et al. [22] has been pivotal in improving the performance of language models. Additionally, techniques such as generative pretraining followed by discriminative fine-tuning, as proposed by Radford et al. [21], have been instrumental in achieving significant improvements in natural language understanding benchmarks. Furthermore, the GPT-4 [23], PaLM [24], LLaMA [25], Qwen [26], and Baichuan [27] have demonstrated the effectiveness of large-scale language models in various domains, including vertical domains and multilingual settings.

The fine-tuning of LLMs for urban-related tasks enables accurate and efficient analysis of urban renewal, urbanization development, and urban planning. Fine-tuning is the prevalent paradigm for using pre-trained language models (LM) to perform downstream tasks, which requires updating all the parameters of the LM [19][28]. GPT-3 uses manually designed prompts to adapt its generation for different tasks, and this framework is termed in-context learning [29]. Sun et al. use keyword prompts to control the sentiment or topic of the generated sentence [30]. Prompt engineering has been explored in natural language understanding models like BERT and RoBERTa [31][32]. Shin et al. use auto-prompt to search for a sequence of discrete trigger words [33]. Prefix-tuning [34] and P-Tuning v2 [35] have shown that optimized prompt tuning can be comparable to fine-tuning across different model scales and language understanding tasks, providing alternative approaches to model adaptation. Besides, techniques like LoRA [36], QLoRA [37], and LongLoRA [38] have addressed the challenges posed by large-scale models, reducing trainable parameters and improving computational efficiency in fine-tuning processes. Adapter tuning [39] has also demonstrated parameter-efficient transfer learning capabilities, while continuous pretraining with longer sequences [40] has shown promise in enhancing long-context capabilities.

## 3. Large Language Model and Fine-tuning Theory

This section introduces ChatGLM and fine-tuning methods used in this study, including Prefix, and LoRA tuning techniques.

### 3.1 ChatGLM

ChatGLM is an open-source dialogue language model based on GLM architecture [18], with 6.2 billion parameters, developed by KEG Laboratory of Tsinghua University and Zhipu AI. The model continues to pre-train texts and codes based on GLM-130B, and achieves human intention alignment through supervised fine-tuning techniques. It can write copy, extract information, role play, question-answering, dialogue, etc.

Regarding model structure, it combines the advantages of two mainstream pretraining model architectures, GPT [21] and BERT [19], enabling the model to utilize contextual information better, thereby improving the accuracy and coherence of generated text. The principles of the ChatGLM mainly include the following points:

(1) Large-scale unsupervised pretraining: ChatGLM uses large-scale internet corpus for unsupervised pretraining. In this way, the model can learn general text features in massive data, thus having strong language comprehension and generation capabilities.

(2) Masked Language Modeling: ChatGLM uses Masked Language Modeling (MLM) technology during the pretraining process. This technology can improve the model's sensitivity to potential information in the text, helping to improve generation quality and accuracy.

(3) Supervised fine-tuning: To adapt to multiple tasks and achieve alignment of human intentions, ChatGLM uses supervised fine-tuning technology. The model can better complete the target task by fine-tuning the pre-trained model on the datasets of different tasks.

The pretraining and supervised fine-tuning methods of ChatGLM enable it to use specific task datasets for further fine-tuning training, making it suitable for completing knowledge QA tasks in urban renewal.

### 3.2 Prefix Tuning

Prefix Tuning is a technique used in fine-tuning LLM parameters, the core idea is to construct a set of task-related virtual tokens as a prefix before inputting the token [34]. The main advantage of this method is that it can optimize the model's performance by adjusting the parameters of the prefix without changing the pre-trained large model (PLM).

Different models require different prefixes. For autoregressive models, a prefix can be added to the front of the sentence to obtain z = [Prefix; x; y], where x and y represent the input context and output context (question and answer in urban renewal QA dataset in this study), respectively. During training, only the parameters of the Prefix layer are updated, while the original parameters in the PLM remain fixed. At the same time, to prevent unstable training and performance degradation caused by directly updating the parameters of the Prefix layer, a multilayer perceptron (MLP) structure is added in front of the Prefix layer [34], as shown and referenced in Figure 1. After training, only the parameters of the Prefix layer are retained. In this way, the Prefix layer can influence the PLM during the process of learning context, which guides the generation of context and makes it better adapted to new tasks.

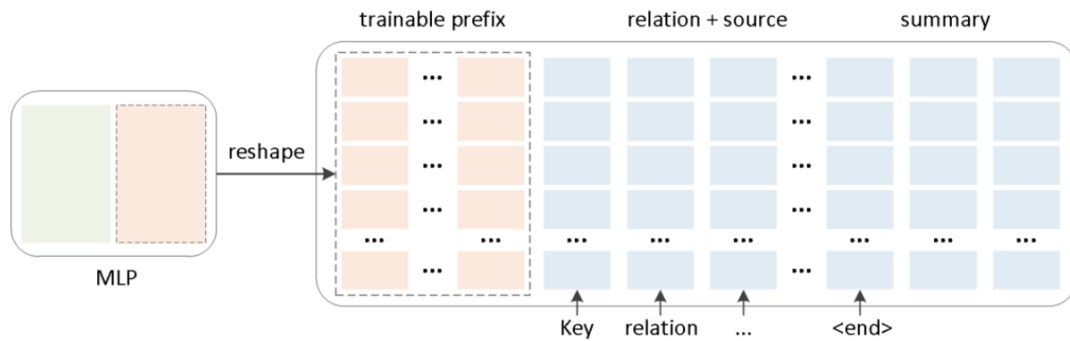

Figure 1. Schematic Diagram of Prefix[1].

**3.3 LoRA**

Low-rank adaptation (LoRA) is a parameter-efficient fine-tuning method, which assumes that the model's parameters change during task adaptation is low-rank [36]. Therefore, the parameter change can be simulated through low-rank decomposition, allowing for the indirect training of LLMs with minimal parameter requirements. The main advantage of this approach is that it can significantly reduce the number of parameters required during fine-tuning, thereby alleviating the burden on computational resources.

It injects a trainable layer (called rank decomposition matrix) into each transformer block, which adds a dimension reduction matrix A and a dimension expansion matrix B next to the linear layer of the model [36], as shown and referenced in Figure 2. During training, the original parameters of the PLM are frozen, and only the parameters of matrices A and B, which are related to urban renewal tasks in this study, are updated. Among them, A reduces the data from d dimensions to r dimensions, where r is the rank of LoRA, an important hyperparameter; B expands the data from r dimensions to d dimensions, and the parameters of part B are initialized to 0. After the model training is complete, the parameters of the A+B need to be merged with the parameters of the PLM for inference. In this way, it can preserve the PLM's powerful capabilities and adapt well to new tasks.

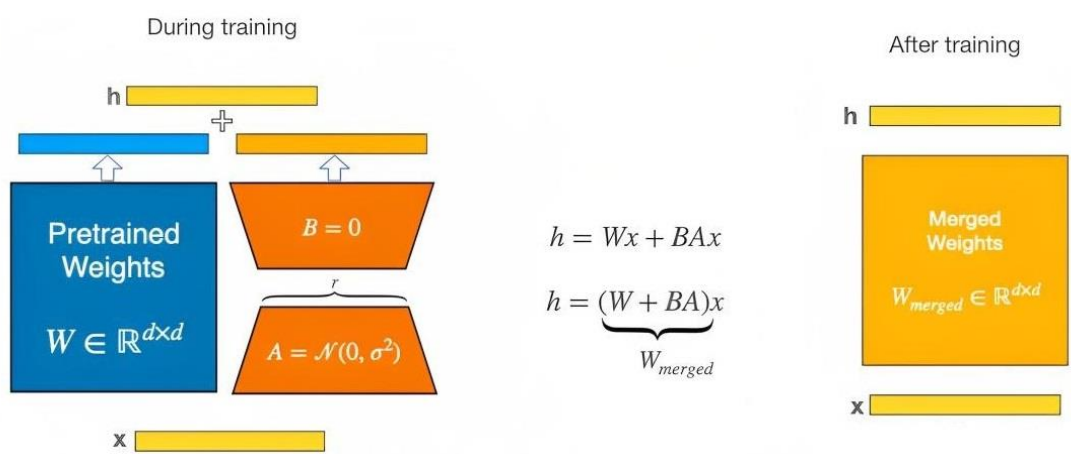

Figure 2. Schematic Diagram of LoRA[2].

---

[1] Figure 1 is cited and modified from reference [34].
[2] Figure 2 is cited and modified from reference [36].

## 4. Building Urban Renewal Large Language Model

This section details the construction methods of urban renewal LLM based on ChatGLM, which includes the following subsections: data preparation, model setup, and performance evaluation.

**4.1 Data Preparation**

This study is based on scientific literature corpora of urban renewal, using the ChatGLM to generate the QA dataset automatically. The specific steps are as follows:

(1) Collect about 70-80 scientific articles related to urban renewal as the source of corpus data;

(2) For each scientific document in pdf format, automatically extract the text using Python's toolkit pdfminer, and perform preliminary preprocessing on the text, such as removing chapter titles, figure titles, table titles, special characters, etc., retaining only the plain text of each paragraph, and further manually select high-quality and appropriate paragraph texts (such as abstracts, introductions, and conclusions) as corpus data;

(3) Deploy the ChatGLM in an API manner on the local server. By writing an appropriate prompt, each piece of processed textual corpus is input to ChatGLM one by one through API calls, guiding the model to automatically generate and return several QA data in a self-instruct manner [41] based on the prompt and given text, including each specific question and corresponding answer about urban renewal.

(4) Further processing these generated QA data, deleting duplicate and erroneous QA pairs, and adding appropriate instruction words constitutes a complete and high-quality QA instruction dataset.

**4.2 Model Setup**

To obtain the LLM for urban renewal, this study uses the QA instruction dataset described in Section 4.1 to conduct fine-tuning training for the ChatGLM, including the following methods: Prefix tuning, LoRA, and Joint fine-tuning.

**4.2.1 Prefix Fine-tuning**

This study first uses the Prefix tuning method to fine-tune the ChatGLM. By inputting the prompt and question as prefix into the model and having the model predict the subsequent answer, we can make preliminary adjustments to the model's adaptability in urban renewal. The specific steps are as follows:

(1) Divide the generated QA instruction dataset into 1085 training and 275 test samples.

(2) Use the training samples to perform Prefix tuning training on the ChatGLM and save the Prefix parameter weights after training.

(3) Use the test samples to evaluate the text generation performance of the ChatGLM after fine-tuning with Prefix tuning, and optimize the model hyperparameters. The optimal model hyperparameter settings are as follows: pre_seq_len=128, learning_rate=2e-2, batch size=16, epochs=5, max_source_length=128, max_target_len=256.

**4.2.2 LoRA Fine-tuning**

Similarly, the LoRA method will be used to fine-tune the ChatGLM and optimize its performance by introducing specific knowledge of urban renewal. The specific steps are as follows:

(1) Divide the generated QA instruction dataset into 1085 training and 275 test samples (as above).

(2) Use the training samples to fine-tune the ChatGLM with LoRA and save the LoRA parameter weights after training.

(3) The text generation performance of the ChatGLM after LoRA fine-tuning is evaluated using the test samples, and the model hyperparameters are optimized and adjusted. The optimal model hyperparameter settings are as follows: lora_r=8, lora_alpha=32,

lora_dropout=0.05, learning_rate=1e-4, batchsize=16, epochs=10, max_source_length=128, max_target_len=256.

**4.2.3 Joint Fine-tuning**

Examining the results of Prefix and LoRA tuning separately, we hypothesized that jointly fine-tuning the ChatGLM may improve the model performance. The specific steps are as follows:

(1) Combine the optimal Prefix parameter weights described in Section 4.2.1 with ChatGLM and save them as the ChatGLM-Prefix model;

(2) Based on the same training set, the LoRA method is used to fine-tune the ChatGLM-Prefix model further. The model hyperparameters are set to the same values as in Section 4.2.2 to ensure comparability.

(3) Use the test samples to evaluate the text generation performance of the model after joint fine-tuning and compare it with the methods described in Sections 4.2.1 and 4.2.2 to verify the superiority of this method.

(4) Before inferring, the ChatGLM-Prefix model is combined with the LoRA parameter weights to obtain the ChatGLM-Prefix-LoRA model, which can be used to complete knowledge QA tasks in urban renewal.

**4.3 Performance Evaluation**

This study uses the Bleu and Rouge metrics to evaluate the performance of the fine-tuned urban renewal LLM.

(1) Bleu: The Bleu metric is an evaluation method based on precision for measuring the quality of text generation. This method counts the number of matched N-grams between the generated text and the reference text. The 1-gram comparison compares each word, while the 2-gram comparison compares each word pair, and so on, regardless of word order. This study uses the Bleu-4 metric, which, when weighted, calculates the accumulated 4-gram. The higher the number of matches, the higher the Bleu-4 score, indicating the better the quality of the generated text.

(2) Rouge: The Rouge metric is an evaluation method based on recall, which is also used to measure the quality of text generation. Like Bleu, it evaluates the generation quality by counting the number of overlapping N-grams between the generated text and the reference text. The Rouge metrics used in this study include Rouge-1, Rouge-2, and Rouge-L. The higher the Rouge score, the better the quality of the text generation.

**5. Experiment and Analysis**

**5.1 Instruction Dataset Generation**

According to the method described in Section 4.1, we can obtain text corpus data related to urban renewal. The prompt written to guide the ChatGLM to automatically generate QA data is as follow:

> **Prompt**:
> *You are now a research expert, your task is to generate several QA datasets based on the following text, including the questions and the answers which are appropriate. The results should output in the format like:*
> *Question: [ ]*
> *Answer: [ ]*
> *Question: [ ]*
> *Answer: [ ]*
> *The text: ```{text}```*

Wherein, {text} is a piece of the processed text corpus to be input into the model. This method guides the ChatGLM to generate QA data on urban renewal topics automatically in the form of self-instruct and zero-shot. By further reviewing and processing the generated questions and answers and adding appropriate instruction words, a high-quality dataset for fine-tuning the LLM can be obtained. An example of a text corpus and generated QA data is shown below:

**Text corpus:**
*Urban renewal planning and development are vital to enhancing the living quality of city residents. However, such improvement activities are often expensive, time-consuming, and need more standardization. The convergence of GIS technologies, social big data and artificial intelligence (AI) solutions has created unprecedented opportunities for comprehensive digital planning and analysis in urban renewable development. However, fast interdisciplinary development imposes challenges because information and digital solutions are piecewise and need more knowledge fusions, integrations, evaluation standard, systematic analysis and methodology. To address these challenges, we proposed an Urban Meta Analysis (UMA) framework with data modeling, and mathematical analysis models supported by the Municipal Urban Development Index (MUDI) system.*

**Generated QA data:**
*Question: What is the purpose of urban renewal planning and development?*
*Answer: The purpose of urban renewal planning and development is to enhance the living quality of city residents.*

*Question: What are the challenges of urban renewal planning and development?*
*Answer: The challenges of urban renewal planning and development are expensive, time-consuming, and need more standardization. The convergence of GIS technologies, social big data, and artificial intelligence (AI) solutions has created unprecedented opportunities for comprehensive digital planning and analysis in urban renewable development, but fast interdisciplinary development imposes these challenges.*

*Question: What is the proposed solution to these challenges?*
*Answer: The proposed solution to these challenges is the use of the Urban Meta Analysis (UMA) framework with data modeling, and mathematical analysis models supported by the Municipal Urban Development Index (MUDI) system.*

**5.2 Model Fine-tuning Results**

According to the methods described in Sections 4.2.1 and 4.2.2, the ChatGLM is trained using 1085 training samples with Prefix fine-tuning and LoRA fine-tuning, respectively. The trainable parameters and performance evaluation results of the two methods on 275 test samples are shown in Table 1. At the same time, to compare the performance before and after model fine-tuning, the ChatGLM without fine-tuning is also evaluated using the same test samples.

Table 1. Parameters and performance evaluation results of fine-tuned ChatGLM model.

| Model | Total Parameters | Trainable Parameters | Bleu-4 | Rouge-1 | Rouge-2 | Rouge-l |
|---|---|---|---|---|---|---|
| ChatGLM | 5954.35M | - | 25.44% | 50.69% | 28.32% | 45.27% |
| ChatGLM-Prefix | 5957.85M | 3.50M | 41.74% | 52.06% | 33.43% | 45.72% |
| ChatGLM-LoRA | 5956.21M | 1.86M | 38.53% | 61.53% | 41.69% | 57.87% |

As shown in Table 1, compared to the ChatGLM without any fine-tuning, the models trained using the urban renewal instruction dataset with Prefix and LoRA fine-tuning have significantly improved scores on the Bleu and Rouge metrics on the test samples, with only small parameters added. In addition, the ChatGLM-Prefix model has a higher Bleu score of about 3% compared to the ChatGLM-LoRA model, but at the same time, it has a lower Rouge score of about 10%. Overall, the fine-tuning of ChatGLM can improve performance in urban renewal knowledge QA tasks.

### 5.3 Model Joint Fine-tuning Results

As described in Section 4.2.3, the ChatGLM is trained using a joint fine-tuning process to the hypothesized advantages of combining the Prefix and LoRA fine-tuning methods, which can further optimize the model's performance on the urban renewal knowledge QA tasks. The trainable parameters and performance evaluation results of the joint fine-tuning method on the test samples are shown in Table 2.

Table 2. Parameters and performance evaluation results of joint fine-tuned ChatGLM model.

| Model | Total Parameters | Trainable Parameters | Bleu-4 | Rouge-1 | Rouge-2 | Rouge-l |
|---|---|---|---|---|---|---|
| ChatGLM-Prefix-LoRA | 5959.71M | 5.36M | 43.78% | 64.05% | 46.84% | 62.63% |

By comparing Table 1 and Table 2, it can be found that the ChatGLM-Prefix-LoRA model, after the joint fine-tuning training, can further improve the Bleu and Rouge metrics scores on the test samples. Although its trainable parameters sum up to those of Prefix and LoRA, it still only accounts for a small part of the model's total parameters, about 0.0899%, which can be ignored. Specifically, compared to the ChatGLM-LoRA model, this model can increase the Bleu and Rouge scores by about 5%, respectively; compared to the ChatGLM-Prefix model, this model can increase the Bleu score by about 2%, and the Rouge score by about 10%-15%; compared to the ChatGLM without any fine-tuning, this model can increase the Bleu and Rouge scores by about 15%-20% respectively. The experimental results prove the effectiveness and superiority of the joint fine-tuning training method using Prefix and LoRA for ChatGLM on the urban renewal knowledge QA tasks.

### 6. Discussion

**6.1 Model Joint Fine-tuning Analysis**

The joint fine-tuning method proposed in this study achieves good results in constructing urban renewal LLM and knowledge QA tasks. By guiding the ChatGLM to automatically generate QA data in a self-instructed manner based on prompt words and given text, it can quickly generate a dataset in the urban renewal domain and provide data support for subsequent fine-tuning training of ChatGLM. By joint fine-tuning training with the Prefix and LoRA methods, the model's generation ability on urban renewal knowledge QA task can be significantly improved. Based on the results, we reasoned that the proposed new method takes full advantage of the Prefix method to guide the model to generate the relevant texts and at the same time, introduce specific information indicators to make the LoRA model more effective, thereby improving the quality and accuracy of the model's text generation.

Firstly, the Prefix method can improve the accuracy of generated text while maintaining the original text structure. This is because the Prefix method can leverage contextual information in the original text, which helps the model generate more consistent text with the reference answer. Secondly, the LoRA method enables the model to generate higher-quality text under unsupervised conditions through self-supervised learning. The LoRA method helps improve the model's generalization ability on unknown data, thereby enhancing the professionalism of generated text. Finally, jointly using the two methods for fine-tuning training can fully leverage their respective advantages, resulting in improved accuracy and coherence of generated text in urban renewal.

In addition, compared to the LoRA fine-tuning method, although the joint fine-tuning method requires more trainable parameters, it only increases by less than 0.1% of the model's total parameters. Still, it can bring about a 5% performance improvement. However, suppose only LoRA fine-tuning is used, and the trainable parameters is increased to the same scale as joint fine-tuning. The model performance remains almost unchanged (as shown in Table 1), which cannot be compared to a joint fine-tuning model. The experimental results demonstrate that the optimization effect of the joint fine-tuning method is very impressive while still significantly reducing the burden of computing resources.

**6.2 Limitations and Future Work**

This study has achieved specific results in constructing LLM for urban renewal, but there are still limitations. Future research works are outlined as follows:

(1) Dataset size: This study's urban renewal QA dataset is relatively small. Additional scientific literature in the field of urban renewal can be collected to expand the sources of textual corpus data and generate more QA datasets, to further improve the model performance in knowledge QA tasks. Besides, there may be some text correlation between QA dataset generated from the same literature corpus. To reduce the possibility of text duplication between the training and test samples, independent test set generated from additional literature corpora can be used for testing to further verify the model's generalization capability.

(2) Training samples and model parameters: Further research can be conducted of the relationship between the training samples and the model parameters. For example, for fixed number of model parameters, studying model performances for different training sample size. Conversely, for fixed training sample size, studying model performances for different model parameters.

(3) Fine-tuning methods: This study explores the application of Prefix, LoRA, and joint fine-tuning methods in constructing urban renewal LLM. In the future, joint fine-tuning method of LoRA followed by Prefix can also be explored to analyze the impact of different fine-tuning sequences on model performance. At the same time, the superiority of the joint fine-tuning method can be further demonstrated by comparing other effective fine-tuning methods, such as P-Tuning v2 [35], Adapter [39], etc.

(4) Model selection: This study uses ChatGLM to construct the urban renewal LLM. In the future, other open-source models, such as LLaMA [25], Qwen [26], Baichuan [27], etc. can be tried to demonstrate the universality of joint fine-tuning method for LLMs fully.

In summary, larger scale of the dataset, more suitable parameters, more models and fine-tuning methods can be applied to improve the application effect of the urban renewal LLM, and to fully demonstrate the capabilities and universality of the proposed method in this study.

**7. Conclusion**

This study aims to construct LLM for urban renewal, focusing on improving the model's performance in knowledge QA task and the quality of text generation. Based on the ChatGLM, we automatically generate the QA dataset using the self-instruct method using urban renewal scientific literature corpora. We then further fine-tuned the model jointly using the Prefix and LoRA fine-tuning methods. We successfully construct the urban renewal LLM. The experimental results show that the proposed joint fine-tuning method can significantly improve the performance of LLM in QA task.

The research study covers the application methods, model constructions, and performance evaluations of LLM in the field of urban renewal, with a view of providing strong support for practical applications. Firstly, LLMs can be applied to consulting services in urban renewal, providing project suggestions and planning for governments, businesses, and individuals. Secondly, intelligent customer service can provide users with query services for urban renewal policies, project progress, and other relevant information. In addition, LLMs can also assist urban planners and designers in designing and evaluating urban renewal projects, ensuring the sustainability and social benefits of the projects.

In terms of future research directions, it is possible to further optimize the model structure, expand to other fields, explore multimodal data applications, and focus on social and environmental impacts.

**Acknowledgments**

This research was supported by Department of Automation, AI for Earth Laboratory, AI for Medicine Laboratory, Cross-strait Research Institute, Tsinghua University. We express our appreciation for our colleagues Linping Deng, Kun Li, Haoming Yang, Gang Yin in the laboratory for their support of data verification and computing preparation.**References**

[1] Zhou Y, Li X, Asrar, Ghassem R, Steven J, etc. A global record of annual urban dynamics (1992-2013) from nighttime lights. Remote Sensing of Environment: An Interdisciplinary Journal, 2018, 219.

[2] Gong P. Remote sensing of environmental change over China: A review. Chinese Science Bulletin 57, 2012: 2793-2801.

[3] Angel S, Parent J, Civco D L, et al. The dimensions of global urban expansion: Estimates and projections for all countries, 2000–2050. Progress in Planning, 2011, 75(2): 53-107.DOI:10.1016/j.progress.2011.04.001.

[4] Wang L, et al. China's urban expansion from 1990 to 2010 determined with satellite remote sensing. Chinese Science Bulletin 57, 2012: 2802-2812.

[5] Gong P, Wang J, Yu L, et al. Finer resolution observation and monitoring of global land cover: First mapping results with Landsat TM and ETM+ data. International Journal of Remote Sensing, 2013, 34(7): 48.DOI:10.1080/01431161.2012.748992.

[6] Chen J, Chen J, Liao A, et al. Global land cover mapping at 30 m resolution: A POK-based operational approach. Isprs Journal of Photogrammetry & Remote Sensing, 2015, 103(may): 7-27.DOI:10.1016/j.isprsjprs.2014.09.002.

[7] Angel S, Chabaeva A, Gitlin L, et al. The dynamics of global urban expansion, 2005.

[8] Bao H, Ming D, Guo Y, Zhang K, Zhou K, et al. DFCNN-Based Semantic Recognition of Urban Functional Zones by Integrating Remote Sensing Data and POI Data. Remote. Sens., 2020, 12, 1088.

[9] Du S, Du S, Liu B, Zhang X. Incorporating DeepLabv3+ and object-based image analysis for semantic segmentation of very high resolution remote sensing images. International Journal of Digital Earth, 2020, 14, 357-378.

[10] Chen B, Xu B, Gong P. Mapping essential urban land use categories (EULUC) using geospatial big data: Progress, challenges, and opportunities. Big Earth Data, 2021, 5, 410 - 441.

[11] Wang X, Chen B, Li X, Zhang Y, Ling X, et al. Grid-Based Essential Urban Land Use Classification: A Data and Model Driven Mapping Framework in Xiamen City. Remote Sens, 2022, 14, 6143.

[12] Li X, Yeh G. Modelling sustainable urban development by the integration of constrained cellular automata and GIS. International Journal of Geographical Information Science, 2000.

[13] Li X, Liu X, Gong, P. Integrating ensemble-urban cellular automata model with an uncertainty map to improve the performance of a single model. International Journal of Geographical Information Science, 2015, 29, 762 - 785.

[14] Li Z, He S, Su S, et al. Public Services Equalization in Urbanizing China: Indicators, Spatiotemporal Dynamics and Implications on Regional Economic Disparities. Social Indicators Research, 2020(1).DOI:10.1007/s11205-020-02405-9.

[15] Pendlebury J, Porfyriou H. Heritage, urban regeneration and place-making. Journal of Urban Design, 2017, 22(4):429-432.DOI:info:doi/10.1080/13574809.2017.1326712.

[16] Landa F. Actual changes in system of urban planning in post-socialist city: the case of Prague. Journal of Architecture and Urbanism, 2016, 40(4):303-310.DOI:10.3846/20297955.2016.1246986..

[17] Hartmann T, Jehling M. From diversity to justice – Unraveling pluralistic rationalities in urban design.Cities, 2018, 91.DOI:10.1016/j.cities.2018.02.009.